\documentclass[conference]{IEEEtran}
\usepackage[english]{babel}
\usepackage[latin1]{inputenc}
\usepackage[T1]{fontenc}
\usepackage{epsfig}
\IEEEoverridecommandlockouts
\usepackage{amsmath}
\usepackage{amssymb}
\usepackage{amsthm}
\usepackage{cite}
\usepackage{subfig}
\usepackage{tabularx}
\usepackage{epstopdf}
\usepackage{multirow}
\usepackage{caption}
\usepackage{subcaption}
\usepackage{mathtools}
\usepackage{xfrac}
\usepackage{units}
\usepackage{xcolor}
\usepackage[hyphens]{url}
\usepackage[export]{adjustbox}
\newcommand{\beql}[1]{\begin{equation}\label{#1}}
\newcommand{\eeq}{\end{equation}}

\newcommand{\be}{\begin{equation}}
\newcommand{\ee}{\end{equation}}
\newcommand{\ba}{\begin{array}}
\newcommand{\ea}{\end{array}}
\usepackage{booktabs}
\usepackage{acronym}
\acrodef{ML}[ML]{Machine Learning}
\acrodef{SOTA}[SOTA]{State-Of-The-Art}
\acrodef{MAC}[MAC]{Multiply-Accumulate}
\acrodef{QoS}[QoS]{Quality-of-Service}
\acrodef{CNN}[CNN]{Convolutional Neural Network}
\acrodef{RNN}[RNN]{Recurrent Neural Network}
\acrodef{TPU}[TPU]{Tensor Processing Unit}
\acrodef{AIMC}[AIMC]{Analog In-Memory Computing}
\acrodef{MVM}[MVM]{Matrix-Vector Multiplication}
\acrodef{PCM}[PCM]{Phase Change Memory}
\acrodef{FC}[FC]{Fully Connected}
\acrodef{CONV}[CONV]{Convolutional}
\acrodef{PU}[PU]{Processing Unit}
\acrodef{ADC}[ADC]{Analog-to-Digital Converter}
\acrodef{IMC}[IMC]{In-Memory Computing}
\acrodef{WL}[WL]{Word-Line}
\acrodef{BL}[BL]{Bit-Line}
\acrodef{DNN}[DNN]{Deep Neural Network}
\acrodef{RRAM}[RRAM]{Resistive Random-Access Memory}
\acrodef{SRAM}[SRAM]{Static Random-Access Memory}
\acrodef{DAC}[DAC]{Digital-to-Analog Converter}
\acrodef{ADC}[ADC]{Analog-to-Digital Converter}
\acrodef{HWA}[HWA]{Hardware-Aware}
\acrodef{SGD}[SGD]{Stochastic Gradient Descent}
\acrodef{NVM}[NVM]{Non-Volatile Memory}
\acrodef{FLMS}[FLMS]{First-Last Mapping Strategy}
\acrodef{LBMS}[LBMS]{Layer-Based Mapping Strategy}
\acrodef{FP}[FP]{Floating-Point}
\acrodef{DL}[DL]{Deep Learning}
\acrodef{TL}[TL]{Transfer Learning}
\acrodef{LLM}[LLM]{Large Language Model}
\acrodef{NN}[NN]{Neural Network}
\acrodef{ViT}[ViT]{Vision Transformer}
\acrodef{NLP}[NLP]{Natural Language Processing}
\begin{document}

\title{Assessing the Performance of Analog Training for Transfer Learning}

\author{\IEEEauthorblockN{Omobayode Fagbohungbe\IEEEauthorrefmark{1}\IEEEauthorrefmark{3}, Corey Lammie\IEEEauthorrefmark{2}, Malte J. Rasch\IEEEauthorrefmark{1}, Takashi Ando\IEEEauthorrefmark{1}, Tayfun Gokmen\IEEEauthorrefmark{1}, Vijay Narayanan\IEEEauthorrefmark{1}}
\IEEEauthorblockA{\IEEEauthorrefmark{1}IBM Research - Yorktown Heights, NY USA
\IEEEauthorrefmark{2}IBM Research Europe, 8803 R\"{u}schlikon, Switzerland\\
\IEEEauthorrefmark{3}Email: omobayode.fagbohungbe@ibm.com}
}
\maketitle
\begin{abstract}
Analog in-memory computing is a next-generation computing paradigm that promises fast, parallel, and energy-efficient deep learning training and transfer learning (TL). However, achieving this promise has remained elusive due to a lack of suitable training algorithms. Analog memory devices exhibit asymmetric and non-linear switching behavior in addition to device-to-device variation, meaning that most, if not all, of the current off-the-shelf training algorithms cannot achieve good training outcomes. Also, recently introduced algorithms have enjoyed limited attention, as they require bi-directionally switching devices of unrealistically high symmetry and precision and are highly sensitive. A new algorithm chopped TTv2 (c-TTv2), has been introduced, which leverages the chopped technique to address many of the challenges mentioned above. In this paper, we assess the performance of the c-TTv2 algorithm for analog TL using a Swin-ViT model on a subset of the CIFAR100 dataset. We also investigate the robustness of our algorithm to changes in some device specifications, including weight transfer noise, symmetry point skew, and symmetry point variability.

\end{abstract}

\begin{IEEEkeywords}
Deep Learning, Hardware Implemented Neural Network, Analog Device,  Additive White Gaussian Noise
\end{IEEEkeywords}
\section{Introduction}
\label{sec:Introduction}
\ac{TL}~\cite{tan2018survey} is a training paradigm that leverages insights obtained from one task to accelerate the training of related but less complex tasks. As the process of fine-tuning a foundational model to many downstream tasks of choice is essentially TL\cite{bommasani2021opportunities}, it plays a leading role in the wide application of \acp{LLM} to areas such as education\cite{orenstrakh2024detecting}, medicine\cite{thirunavukarasu2023large}, autonomous driving\cite{chen2024driving}, and finance\cite{li2023large}.
This process can significantly reduce the cost of training large models, as the foundation model can be re-used many times. Furthermore, \ac{TL} makes it possible to train models with high accuracy, even with limited training data\cite{wei2022emergent,zhao2023comparison}.
These benefits have greatly improved the application of \acp{LLM} to tasks such as language modeling\cite{wei2022emergent,yin2023survey}. Additionally, it also offers potential for model deployment at the edge -- particularly in scenarios where model training in the cloud is not possible due to data security and privacy concerns~\cite{yang2021joint,fagbohungbe2021efficient}.

\begin{figure}[!t]
	 \centering
    	 \includegraphics[width=0.45\textwidth]{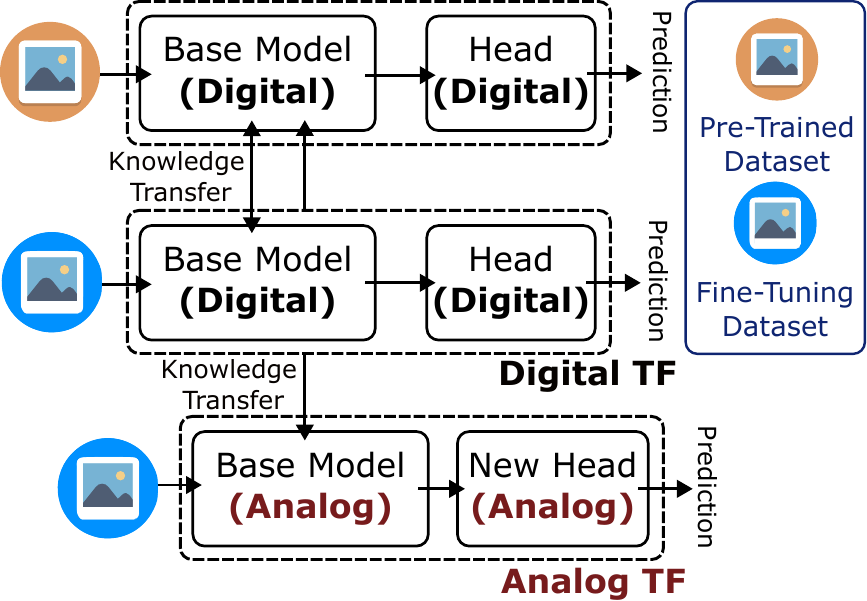}
     	\caption{Key conceptual differences between digital \ac{TL} and our analog \ac{TL} implementation. We obtain the best results when the base model is converted from digital to analog and a new analog head is introduced.}
    \label{fig:digitalandanalgTL}
\end{figure}

However, training \ac{TL} models using conventional hardware is very energy inefficient, particularly at the edge, where devices operate under a strict power budget~\cite{athena2024demonstration}. This inefficiency is due to the von Neumann architecture, creating a physical separation between memory and computing units~\cite{geoburr, jain2022heterogeneous, frenkel2023bottom}. One significant consequence of this separation is a need for continuous data transfer, causing increased power consumption and processing time~\cite{sebastian2019computational}. \ac{AIMC}~\cite{le2023using, geoburr} is a new computing paradigm that represents weight stationary matrices in resistive cross-bar arrays and leverages physical laws such as Kirchhoff's and Ohm's laws to compute matrix-vector multiplications in-memory which has been touted as a possible replacement for current \ac{DL} acceleration hardware~\cite{geoburr,okazaki2022analog}. 

Training \acp{NN} to achieve high-accuracy on \ac{AIMC}-based hardware is challenging. First, the standard Stochastic gradient descent (SGD) training algorithm can not be used with \ac{AIMC} as the computing devices exhibit asymmetric and non-ideal switching behavior~\cite{lee2022impact,rasch2024fast}. Moreover, the asymmetric nature of the memory device also makes gradient accumulation and update of weights represented in the conductances of the memory elements difficult~\cite{kim2022neural}. Conducting some operations, like a full reset to a typical target conductance, is also costly, as the conductance can only be updated in small increments. Device materials also suffer from saturation to minimal and maximum conductance values. Lastly, the inherent presence of device-to-device variation means that an existing or new algorithm that assumes translational invariance cannot be used to train the model~\cite{rasch2024fast, rasch2019training}.

Many efforts have been made to find an optimal algorithm for training \ac{NN} models on \ac{AIMC}-based hardware. The work in~\cite{nandakumar2018mixed} proposed a mixed-precision algorithm that computes the gradient accumulation and weight update stages in digital and the forward and backward passes in analog, sacrificing speed and energy efficiency~\cite{rasch2024fast}.
The authors in~\cite{tayfun2016acceleration} proposed a fast and highly efficient parallel in-memory method to compute the outer-product and weight update operations, leveraging the coincidence
of voltage pulse trains. However, this method requires a bi-directionally switching device of unrealistically
high symmetry and precision, which is difficult to achieve in practice~\cite{tayfun2016acceleration,gokmen2020algorithm,gokmen2021enabling}. The use of the current asymmetric devices with realistic device-to-device variation for gradient accumulation causes devices to drift towards a different conductance value, even in the case when random fluctuations with zero mean are accumulated~\cite{rasch2024fast,gokmen2021enabling}. To solve this problem, the authors in~\cite{gokmen2021enabling,gokmen2020algorithm} proposed various ideas to relax the symmetric and precision requirement, such as the use of separate non-volatile devices to accumulate the weight and gradients and represent pre-determined reference value, the introduction of a low-pass digital filtering stage, etc. 

Despite statically correcting for device-to-device variations on the gradient accumulation and the update stages and achieving good time complexity, the improved training algorithm, known as TTv2, is very challenging to use in practice. The challenge stems from the sensitivity of the training algorithm to even slight changes from programmed reference values to their theoretical values -- it is unlikely that the reference value remains unchanged over a long period of time~\cite{gokmen2021enabling,gong2022deep}. The authors in~\cite{rasch2024fast} proposed a method leveraging a chopped technique~\cite{imperfections1996circuit} to manage any offsets inflicted by an erroneous reference value in the gradient accumulation process by periodic or random sign changes. This new algorithm, called c-TTv2, achieved state-of-the-art results for analog training from scratch on toy models. The details about the c-TTv2 model can be obtained in~\cite{rasch2024fast}.

However, the validation and demonstration of \ac{TL} using the c-TTv2 algorithm has yet to be reported. Hence, this paper aims to demonstrate the use of the c-TTv2 algorithm for TL purposes for \ac{AIMC}-based hardware using a medium-sized \ac{DL} model. We also seek to validate the robustness of the c-TTv2 algorithm to changes in device specifications. Although hardware and software simulation of analog \ac{TL} was performed in \cite{athena2024demonstration}, the TTv2 algorithm algorithm was used. Furthermore, a 3-layer \ac{NN} model and a 4-layer CNN-based model were used for the hardware demonstration and software simulation, respectively. To our knowledge, this is the first paper to simulate training of \ac{AIMC}-based hardware for a \ac{ViT} model.

\section{Methodology}
\label{sec:methods}
In this section, we discuss our method used to validate the effectiveness of the c-TTv2 training algorithm for analog \ac{TL}.
Fig.~\ref{fig:digitalandanalgTL} depicts the block diagram of digital and analog TL, highlighting the differences and similarities between the two approaches. The uniqueness of analog TL stems from using the analog-equivalent base model of the pre-trained model in the fine-tuning phase of the model training. The digital TL differs from the analog TL in that both the pre-training and fine-tuning phases are done in digital.

Data augmentation was performed during the pre-training step. The softmax layer, the last layer of each model, reflects the number of classes in the pre-trained dataset. 
In preparation for the fine-tuning stage, the softmax layer of the pre-trained model was changed to reflect the number of classes of the fine-tuning task. 
Model training was performed for 100 epochs to achieve convergence. The weights and bias of the pre-trained model were then saved. For digital \ac{TL}, the pre-trained model was then fine-tuned using the dataset selected for the fine-tuning phase for a further 100 epochs until convergence was achieved.

To perform analog \ac{TL}, the modified pre-trained model was converted to its analog equivalent, and the fine-tuning process was performed. Due to the difference in precision between analog and digital hardware, it is impossible to program the digital weights to analog conductance values perfectly. To account for this, a weight transfer noise is modeled. This paper defines the weight transfer noise as an additive Gaussian noise of zero mean and unit variance. The mathematical equation defining the relationship between analog weight and pre\-trained weight is
\begin{equation}
W_{noise}=W_{pre\_trained} + \tau \mathbb{N}(0,1),
\end{equation}
where $W_{noise}$ is the analog weight, $W_{pre\_trained}$ is the digital weight, $\tau$ is the noise factor, and $\mathbb{N}(0,1)$ is the Gaussian noise. The noise factor is used to increase or decrease the noise magnitude. A noise factor is a non-zero value, as a noise factor value of zero means that the weight transfer noise is not applied. It should be noted that the weight transfer noise is only added once before the start of the fine-tuning process.
The fine-tuning process for the analog \ac{TL} model was then repeated for different magnitudes of the weight transfer noise to measure its impact on the resulting analog model inference performance. Furthermore, analog and digital model training from scratch was also performed to establish baselines to validate the effectiveness of analog \ac{TL}. These experiments are equivalent to performing the fine-tuning process without the pre-trained weights. 
To investigate the robustness of the c-TTv2 algorithm against selected device specifications, the value of the device specification of interest is increased, and the training process is repeated until convergence is achieved. The learning rate is also varied to select the model that achieves the best model inference performance with the varied device specifications.

We perform simulations using the Swin-ViT~\cite{liu2021swin} model, which has six blocks and 26,591,906 parameters and 26,594,213 parameters for the 2-class task model and 5-class task model respectively. The Swin-ViT\cite{liu2021swin} model belongs to a class of \ac{ViT}~\cite{vaswani2017attention,han2022survey}, which is a class of attention-based architecture models that are widely used in \ac{NLP} due to their computational efficiency and scalability. \ac{ViT} operates by splitting an image into patches and providing the sequence of linear embeddings of these patches (equivalent to tokens) to the attention layers of the transformer. The uniqueness of the Swin-ViT model stems from the usage of a shifted windowing scheme to process patches, bringing greater efficiency by limiting self-attention computation to non-overlapping local windows while also allowing for cross-window connection~\cite{liu2021swin}. 

We use the CIFAR10 dataset for the model pre-training and a subset of the CIFAR100 dataset for the downstream/fine-tuning task~\cite{krizhevsky2009learning}. The CIFAR10 dataset is selected for the pre-training task as it contains more images per class (5000) than the CIFAR100 dataset (500). \ac{TL} requires the complexity of the fine-tuning task to be less than that of the pre-trained task. Hence, in our simulations, 2-class or 5-class subsets of the CIFAR100 dataset are used in the fine-tuning phase.
 
Simulation is performed using Pytorch and the IBM Analog hardware acceleration kit (aihwkit)~\cite{le2023using}. The aihwkit is an open-source toolkit integrated within Pytorch to enable the simulation of analog crossbar arrays. It is centered around the concept of analog tile, a building block that captures the computation performed on a crossbar array. The toolkit is designed to use customized unit cell configurations and advanced optimization algorithms like tiki-taka algorithms. The toolkit can simulate analog tiles of different device materials such as ReRAM, PCM ECRAM, etc.
We simulate HfOx-based ReRAM devices, as they have many desirable properties, such as non-volatility, energy efficiency, high density, and ability to scale. A detailed insight into the device structure can be found here~\cite{gong2022deep}. We list all relevant hyperparameters used in Table~\ref{tab:ResNet50_se}.

\begin{table}
\centering
\captionof{table}{Selected hyperparameters of the c-TTv2 algorithm for Analog Training and  Stochastic Gradient descent Algorithm for Digital Training.}
\label{tab:ResNet50_se}
\begin{tabularx}{0.45\textwidth}{>{\hsize=0.6\hsize}X>{\hsize=0.4\hsize\centering\arraybackslash}X}
\toprule \toprule
Hyperparameter Description & Value \\ 
\midrule
\multicolumn{2}{c}{\textbf{Digital Training}} \\ \midrule
Learning rate &  0.01 \\
Batch size & 8 \\
\midrule
\multicolumn{2}{c}{\textbf{Analog Training}} \\
\midrule
Learning rate &0.01 \\
Batch size & 8 \\
Transfer\_every & 1.0 \\
Autogranularity & 10000 \\
Device momentum & 0 \\
Auto\_scale & True \\
In\_chop Probability & 0.10 \\
Units in mbatch & False \\
Forget Buffer & True \\
Auto\_Momentum & 0.99 \\
\bottomrule \bottomrule
\end{tabularx}
\end{table}

\section{RESULTS}
\label{subsec:ExpSetup}
\subsection{{Comparison between Analog TL and Digital TL}}
\begin{figure}[!t]
	 \centering
    	 \includegraphics[width=0.5\textwidth]{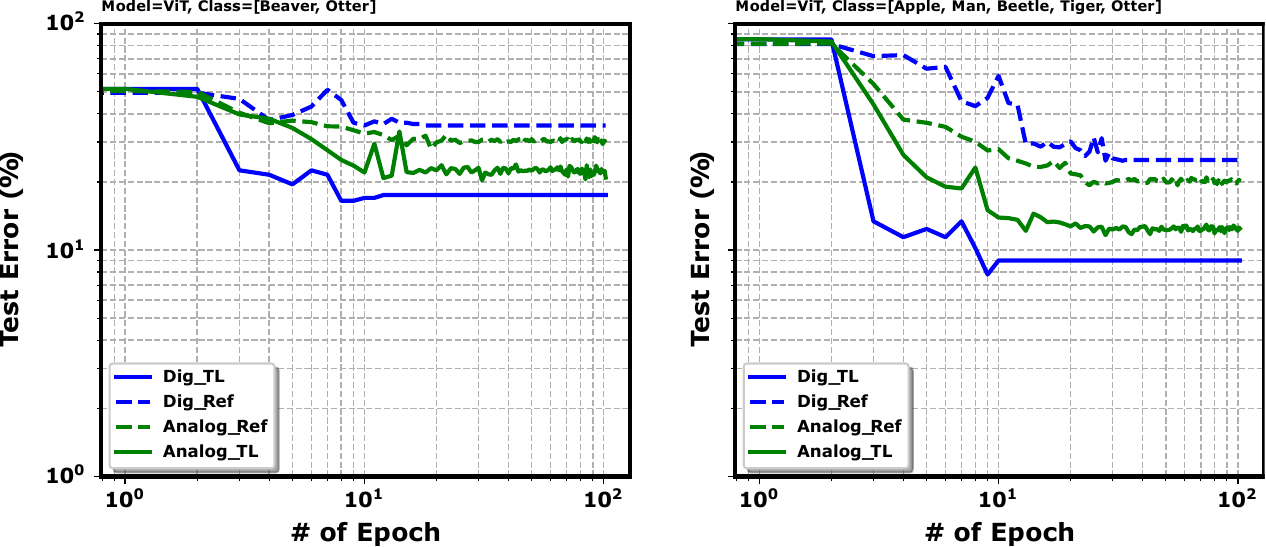}
     	\caption{Training results of the analog and digital Swin-ViT model for TL and regular training (referred to as Ref) performed on the 2-class task (left) and 5-class task (right) using c-TTV2 algorithm}.
    \label{fig:trainingresult}
\end{figure}

Fig.~\ref{fig:trainingresult} shows the test error trace for analog and digital TL for a 2-class and 5-class task for the Swin-ViT model. The figure also shows the test error trace when the models under consideration are trained from scratch. For the \ac{ViT} model, the performance of the analog TL model is better than the results for regular analog training for both fine-tuning tasks. It is also shown that the analog TL model outperforms the digital model's performance. These observations can be explained by the presence of additional noise in the analog model training process. Introducing additional noise during training is equivalent to performing some form of noise injection during the forward pass of the training process. Noise injection during the model training process can be beneficial as it is used for some form of model regularization, which can help improve the model inference performance\cite{fagbohungbe2022impact}. This form of model regularization can be highly beneficial in a TL scenario where the amount of training data is limited. However, the inherent presence of noise in analog model training can be problematic if the training process is sensitive to these noise sources.

For the 2-class and 5-class fine-tuning tasks, digital TL outperforms the analog TL by about 2\%. The performance difference increases with respect to the task complexity. This difference is expected as the increase in task complexity increases the computation's noisy nature, leading to lower analog model inference performance. However, the performance difference is insignificant compared to the earlier algorithm\cite{rasch2024fast}, validating the effectiveness of the c-TTv2 algorithm. 

\label{sec:discussion}
\begin{figure}[!t]
	 \centering
    	 \includegraphics[width=0.45\textwidth]{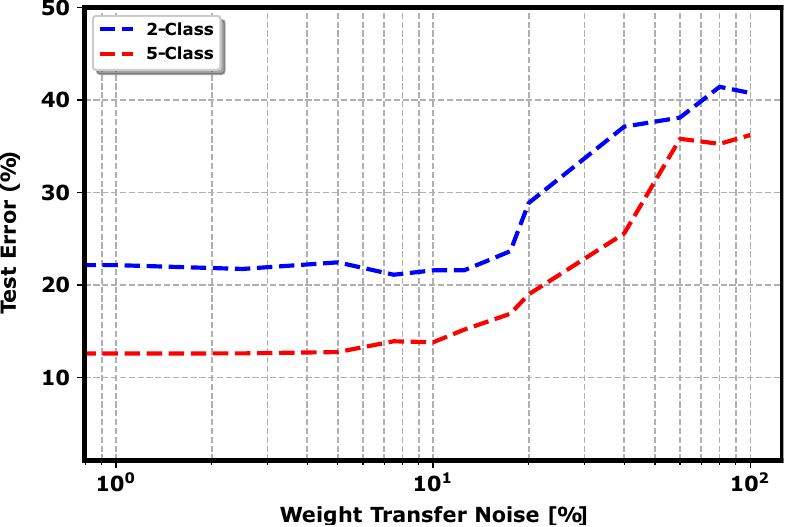}
     	\caption{Relationship between the inference accuracy and weight transfer noise for Swin-ViT analog TL model for the 2-class and 5-class tasks.}
    \label{fig:weightransfernoise}
\end{figure}

\subsection{{Analog TL and Weight Transfer Noise}}
\label{sec:discussion2}
In Fig.~\ref{fig:weightransfernoise}, we determine the effect of the weight transfer noise on analog TL inference performance. For a 2-class task, the test error is mostly constant until the weight transfer noise exceeds $\approx$15\%. For larger noise magnitudes, the degradation in model performance becomes more substantial. Similar behavior is also noticed for the 5-class task, except that the elbow point was $\approx$10\%. These observations imply that the weight transfer noise can negatively affect model inference performance. At weight transfer noise below a certain point (or elbow), called the critical weight transfer noise, the effect of weight transfer noise on the analog TL model inference performance is negligible. The critical weight transfer noise point is affected by the complexity of the task and the model itself. 

\subsection{{Analog TL and Device parameters}}
\label{sec:discussion3}
\begin{figure}[!t]
	 \centering
    	 \includegraphics[width=0.5\textwidth]{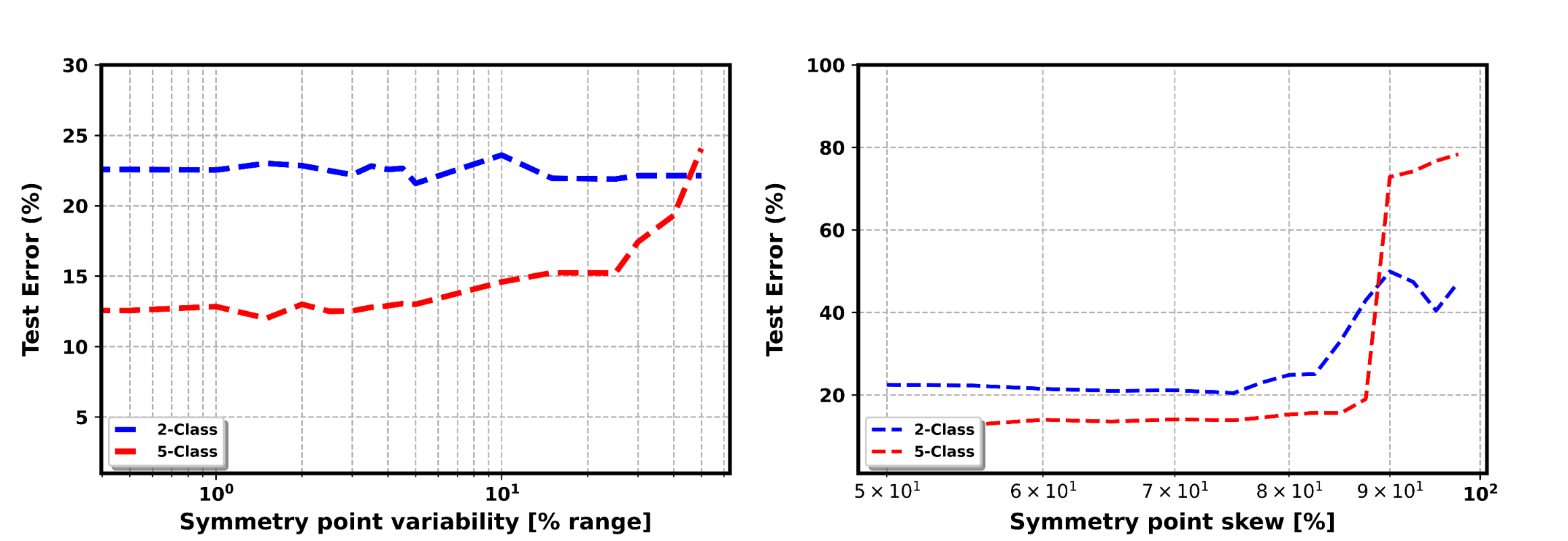}
     	\caption{Relationship between the model inference accuracy and the symmetry point skew for Swin-ViT analog TL model for the 2-class and 5-class tasks trained using the c-TTv2 algorithm.}
    \label{fig:sympointskew}
\end{figure}

\begin{figure}[htbp]
	 \centering
    	 \includegraphics[width=0.5\textwidth]{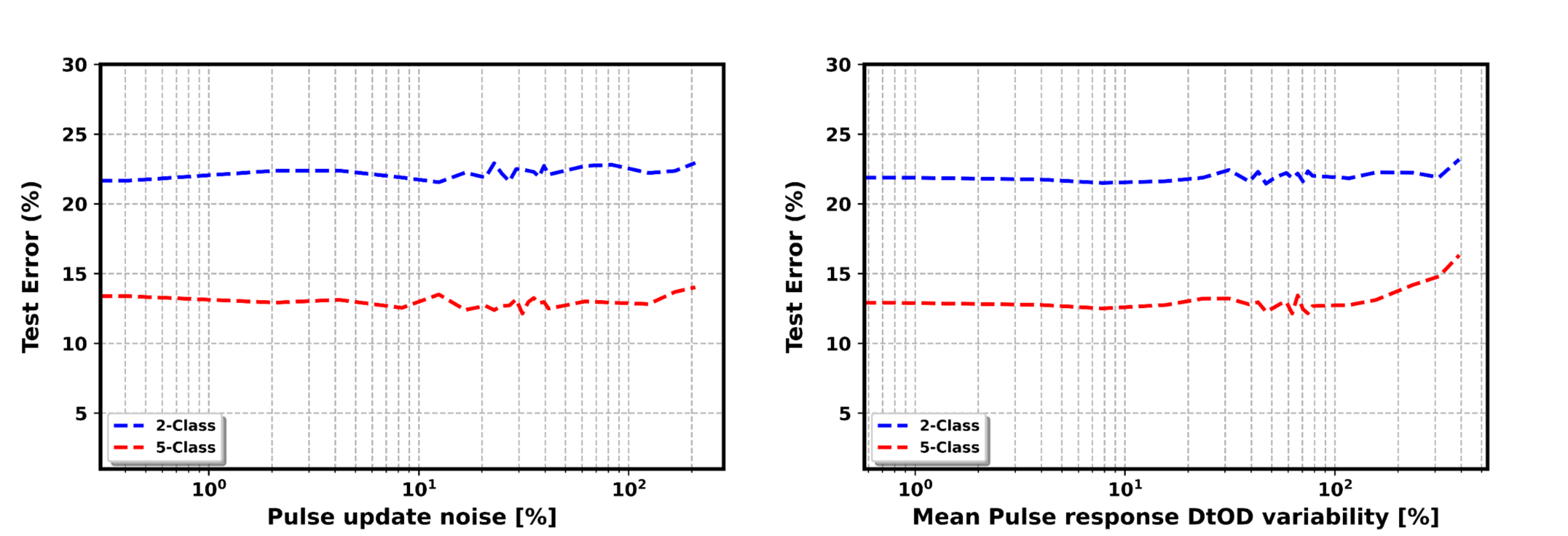}
     	\caption{ Relationship between the model inference accuracy, pulse update noise (left), and mean pulse device-to-device variation for Swin-ViT analog TL model for the 2-class and 5-class tasks trained using the c-TTv2 algorithm.}
    \label{fig:pulse}
\end{figure}

One major requirement of the proposed training algorithm to solve the problem of non-linear switching behavior in ReRAM devices is the need for a bi-directional switching device of unrealistically high symmetry and precision. Improvement in the algorithm further relaxes the requirement for high symmetry, but the need for it still exists. Hence, there is a need to investigate the robustness of the training algorithm in relation to the presence/absence of symmetry for analog TL. The behavior of the ReRAM device for this work, which is modeled using the softbound device model, is used to simulate the ReRAM device used in this work as this model is more representative of the device behavior. Hence, the impact of symmetric point skewness (SpS) and symmetric point variability (SpV) is discussed in this section. The effect of symmetric point skewness on the model inference accuracy is shown in Fig.~\ref{fig:sympointskew}. The device skewness is a measure of how far the maximum weight $W_{max}$ is from the ideal maximum weight value of 1 (or how far $W_{min}$ is from the ideal minimum weight). It is calculated in terms of a percentage (\%),
\begin{equation}
\centering 
SpS=\frac{W_{max}*100}{W_{max}-W_{min}}.
\end{equation}
The symmetric point variability is calculated as a percentage of the weight range of 2 (+1,-1). A 5\% symmetric point variability is equal to a variability value of 0.10. For the 2-class task, the symmetric point variability has minimal impact on the model performance for the symmetric point variability values under consideration, showing the robustness of the training algorithm. For a 5-class task, the analog model inference performance was not affected until a symmetric value of about 10\%. At symmetric point variability beyond this point, the model inference error grows, meaning that the training algorithm has limited robustness. This observation implies that the robustness of the training algorithm to symmetric point variability is affected by the complexity of the task. Fig.~\ref{fig:pulse} investigates the impact of pulse update noise and mean pulse device-to-device variability on model performance. It can be observed that the analog model inference performance remains mostly the same despite the increase in the value of the mean pulse device-to-device (DtoD) variability and the pulse update noise. The same observation can be made for the 2-class and the 5-class fine-tuning tasks, confirming the robustness of the c-TTv2 training algorithm to the noise in the pulse response. 


\section{{Conclusion}}
\label{sec:conclusion}
In this paper, we assessed the performance of the c-TTv2 algorithm on an Analog Swin-ViT model for an analog \ac{TL} task. We performed additional simulations to investigate the robustness of the c-TTv2 training algorithm to changes in the material specifications. Results demonstrated that c-TTv2 is suitable for training the analog TL models as it achieves better results than analog training from scratch and digital training from scratch, as expected. We also demonstrated that the c-TTv2 algorithm achieves competitive results with digital TL. Critically, the c-TTv2 algorithm was very tolerant to changes in device material specification, such as weight transfer noise, symmetry point variability, symmetry point skew, pulse update noise, and mean pulse DtoD variability.
In future work, we aim to assess the performance of the c-TTv2 algorithm using more complex models and tasks such as LLM models, diffusion models, etc. Furthermore, there is a need to benchmark the performance of the training algorithm against the AGAD algorithm\cite{rasch2024fast}. Lastly, there is a need to propose solutions that help narrow the inference performance gap between analog TL and digital TL models.
Our intention for this paper is to lay the initial groundwork required for these future research directions.

\newpage
\bibliographystyle{IEEEtran}
\bibliography{Benchmarking}

\end{document}